\documentclass[a4paper,UKenglish,cleveref, autoref, thm-restate]{lipics-v2019}

\usepackage{multirow}

\title{Evaluating the Effectiveness of Large Language Models in Representing and Understanding Movement Trajectories} 

\titlerunning{LLMs in Representing and Understanding Movement Trajectories} 

\author{Yuhan Ji}{GeoDS Lab, Department of Geography, University of Wisconsin-Madison, USA }{yuhan.ji@wisc.edu}{https://orcid.org/0000-0003-4300-1436}{}

\author{Song Gao}{GeoDS Lab,  Department of Geography, University of Wisconsin-Madison, USA }{song.gao@wisc.edu}{https://orcid.org/0000-0003-4359-6302}{The authors would like to acknowledge the support from the H.I. Romnes Fellowship and the collaborative project between the GeoDS Lab@UW-Madison and Arity.}

\authorrunning{ }

\Copyright{Yuhan Ji and Song Gao} 



\relatedversion{} 

\supplement{}

\nolinenumbers 

\EventEditors{Grant McKenzie, Benjamin Adams, Amy Griffin, and Simon Scheider}
\EventLongTitle{The 16th Conference on
Spatial Information Theory (COSIT 2024)}
\EventShortTitle{COSIT 2024}
\EventAcronym{COSIT2024}
\EventYear{2024}
\EventDate{September 17--20, 2024}
\EventLocation{Québec City, Canada}
\EventLogo{}
\SeriesVolume{X}
\ArticleNo{X}

\begin{document}

\maketitle

\begin{abstract}
This research focuses on assessing the ability of AI foundation models in representing the trajectories of movements. We utilize one of the large language models (LLMs) (i.e., GPT-J) to encode the string format of trajectories and then evaluate the effectiveness of the LLM-based representation for trajectory data analysis. The experiments demonstrate that while the LLM-based embeddings can preserve certain trajectory distance metrics (i.e., the correlation coefficients exceed 0.74 between the Cosine distance derived from GPT-J embeddings and the Hausdorff and Dynamic Time Warping distances), challenges remain in restoring numeric values and retrieving spatial neighbors in movement trajectory analytics. In addition, the LLMs can understand the spatiotemporal dependency contained in trajectories and have good accuracy in location prediction tasks. This research highlights the need for improvement in terms of capturing the nuances and complexities of the underlying geospatial data and integrating domain knowledge to support various GeoAI applications using LLMs. 
\end{abstract}

\section{Introduction}
\label{sec:intro}

The emergence of powerful AI foundation models (e.g., Large Language Models-LLMs, pre-trained vision and multi-modal models) has captured the enthusiasm of researchers, developers, and decision makers in various domains. The development of foundation models (e.g., Generative Pre-trained Transformer, GPT) is still in its early stages towards artificial general intelligence, but the potential society benefits are enormous.  The model pre-training process involves learning a vast amount of data from a wide range of resources, thus being able to adeptly grasp linguistic patterns, comprehend contextual information, and connect knowledge from different domains.
The extensive scale empowers such pre-trained LLMs to be applied directly or transfer to a wide range of cross-domain tasks after minor fine-tuning or few-shot/zero-shot learning, e.g. education \cite{kasneci2023chatgpt}, transportation \cite{zheng2023chatgpt}, healthcare \cite{yang2022large}, GIS skills and geospatial semantics \cite{mooney2023towards,mai2023opportunities}, etc. The tasks can be in the form of question answering, content generation, classification, causal inference, etc. Researchers have explored the use of LLMs to ensure text coherence and consistency in time series data analysis \cite{mai2023opportunities}.
 
Inspired by the preliminary works on GeoAI and movement data analytics~\cite{dodge2020progress,gao2023handbook,janowicz2020geoai}, we are motivated to utilize LLMs for trajectory representation and analysis, which is fundamental to understanding human mobility, urban dynamics, and animal ecology. Trajectory data inherently consists of spatiotemporal sequence data, thus being compatible with LLMs designed for textual sequences. Despite the abundance of trajectory datasets, the analytical framework is usually customized to address specific problems in human movements or animal ecology~\cite{demvsar2021establishing}. Moreover, current models usually rely on subjective decisions on the discretion of the space and time extents at the cost of resolution lower than the documented~\cite{krumm2006real, song2010limits}. While it is possible to extend LLMs to disruptive movement analysis, the degree to which a trajectory can be properly represented and the mobility patterns can be preserved remains to be examined, which will be the focus of this work.

\section{Methodology}
\subsection{Overview}

The work aims to assess the capacity of LLMs in representing and conducting reasoning on available movement trajectory data. 
From longitudinal trajectory data collected from location-based services, we are able to derive two types of trajectories: First is the sequence of stay points where individuals were engaged in activities. Second, is the travel path between two consecutive stay points, referred to as the origin and destination (OD), which records the locations (e.g. GPS fixes) traversed during the trip at a fine spatiotemporal resolution. The obtained trajectories (sequences of coordinates) are then formatted as input into LLMs.

We hypothesize that an LLM can serve as an encoder to learn the high-dimension representation (embedding) of a trajectory composed of coordinates, which can capture contextual information, such as temporal dependency and spatial patterns. Additionally, by incorporating effective prompts and employing fine-tuning or few-shot learning \cite{wang2020generalizing}, the models can potentially leverage the newly acquired knowledge as well as the pre-trained knowledge to perform specific tasks or offer valuable insights. We identify two extensively studied trajectory mining tasks (T1 and T2) for validation and evaluation (See Section \ref{problem}).

\subsection{Dataset}

The GeoLife human movement trajectory dataset~\cite{zheng2010geolife}  is employed to evaluate our proposed evaluation framework. The GeoLife dataset  consists of individual users' outdoor movements with 17,621 trajectories from 182 users over a period from April 2007 to August 2012 using GPS loggers and GPS phones. A GPS trajectory consists of a sequence of time-stamped points of longitude and latitude. Due to the difference in tracking devices, the sampling rates are not the same across users, with majority of users sampled at about every 5 seconds. Since the participants were recruited through rewards, the total duration of an individual's trajectories varies from weeks to years. 
The Python library \textit{scikit-mobility} \cite{JSSv103i04} and \textit{scikit-learn} are used to preprocess the dataset, including noise filtering, trajectory compression, stay points identification, stay points clustering, and user filtering.

\subsection{Foundation Model}
The GPT-J from EleutherAI~\cite{gpt-j} is used in this work to explore the potential application of LLMs in trajectory modeling. The GPT-J project aims to replicate OpenAI's GPT-3 model with a reduced number of parameters while making it open source to the public. Therefore, like GPT-3, GPT-J enables few-shot learning in text analysis and serves specified downstream tasks.

\subsection{Problem statement} \label{problem}

\noindent \textbf{T1: Trajectory distance measures}

\noindent \textbf{Description:} In this task, we focus on the global trajectory distances which consider the overall similarity between a pair of trajectories regarding all the points contained \cite{zheng2011computing}. We compare the distance between two trajectories using two approaches: traditional distance metrics and the distance between embeddings.

\begin{figure}[h]
    \centering
    \includegraphics[width=1\linewidth]{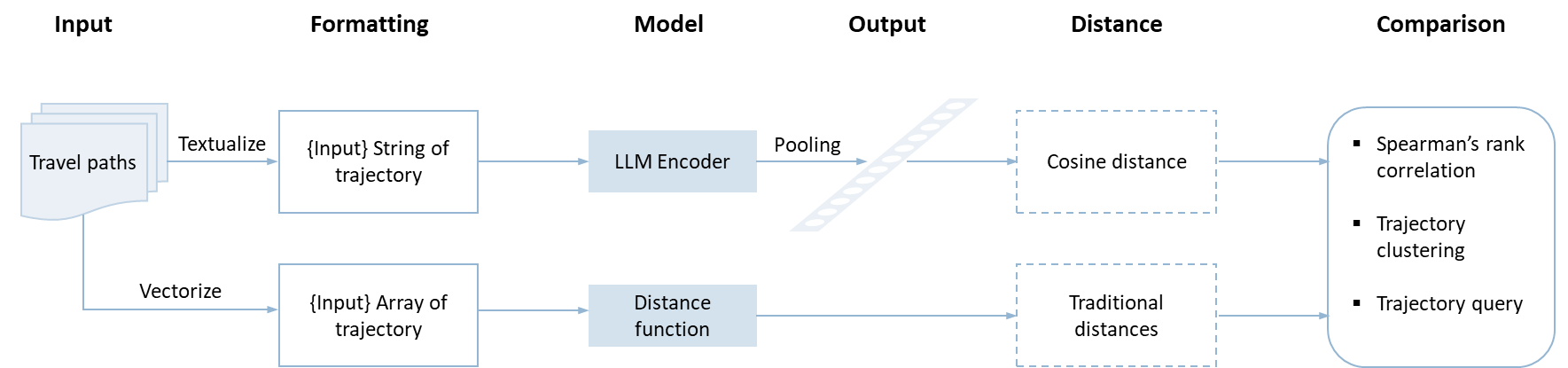}
    \caption{The evaluation workflow of trajectory distance measures}
    \label{fig:distance}
\end{figure}

\noindent \textbf{Method:} The effectiveness of using LLM-based embedding for measuring trajectory distance is assessed as in Figure \ref{fig:distance}.

Given the entire travel trajectories, we first convert it to a string of coordinates, e.g., ``Trajectory: (116.31752, 39.98461), (116.31338, 39.98459), ...,  (116.3087, 39.98455)''. Input the formatted string into an LLM to obtain the token embeddings. Next we generate the trajectory embeddings by an average pooling operation of the token embeddings. The cosine distance is adopted to measure the proximity between two trajectory embeddings. In addition, commonly used traditional trajectory distances including Hausdorff distance, DTW (Dynamic Time Warping), and Longest Common Subsequence (LCSS) are also computed using the original trajectory sequences. 

\noindent \textbf{Evaluation:} Spearman's rank correlation between the distances is computed from two approaches (LLMs vs. traditional) for the same pair of trajectories. In addition, the obtained distance matrices are used for hierarchical trajectory clustering with a fixed number of clusters  and k-nearest trajectory retrieval. The consistency between trajectory clusters of Hausdorff distance and Cosine distance is evaluated by the Rand Index. The percentage of the agreeing pairs of nearest neighbors is computed. 

\noindent \textbf{T2: Contextless destination prediction}

\noindent \textbf{Description:} The aim of the task is to predict the destination given the partial travel path of an ongoing trip. In a contextless data setting, only trajectory coordinates are available without contextual information such as timestamps and place semantics \cite{tsiligkaridis2020personalized}. 

\noindent \textbf{Method:} Given that the destination information is not originally contained in the partial trajectories, a potential approach is to leverage the generation capability of LLMs to produce destinations by fine-tuning a causal language model (CausalLM). The training prompts contain both the partial trajectories and destinations, for instance, ``Trajectory: (116.3248, 40.012), (116.3217, 40.0107), (116.3242, 40.0091), (116.3244, 40.0061) => Destination (116.3262, 40.0002)''. However, during testing, the destinations are omitted from the prompts and need to be completed by the fine-tuned model. In order to build a comparative baseline, we implement the trajectory distribution model proposed in \cite{besse2017destination}. The model predicts the destinations based on the Gaussian Mixture Model clustering of trajectory points.

\begin{figure}[h]
    \centering
    \includegraphics[width=1.0\linewidth]{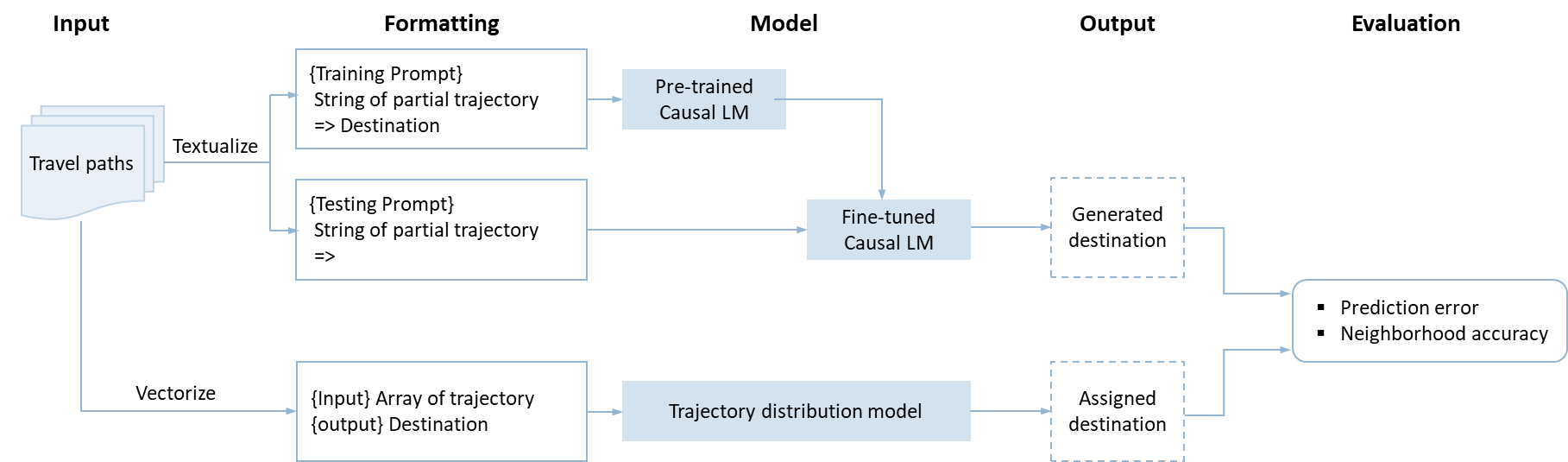}
    \caption{The evaluation workflow of trajectory destination prediction}
    \label{fig:destination}
\end{figure}

\noindent \textbf{Evaluation:} We first need to check whether the output text is valid coordinates (i.e., within the latitude [-90,90] and longitude [-180,180] ranges in degree). If valid, we adopt similar evaluation metrics as presented in \cite{tsiligkaridis2020personalized}. The prediction error is calculated by the haversine distance between the ground truth and predicted destinations. A prediction is classified as accurate if it is within a given distance of the true destination. In addition, because a generative model may generate multiple possible results, we further extend the performance metrics to Error@k and Accuracy@k, which are the mean error of k trials and the fraction of times the destination is correctly predicted among the top k candidates.






\section{Results} \label{results}
\subsection{Trajectory distance measure results} \label{results}

The GeoLife trajectories of medium length (based on percentile) are selected for experimental evaluation. We compute the pairwise trajectory distances according to the details described in Section \ref{problem}. All the distance measures are parameter-free except LCSS, which requires a spatial threshold $\epsilon$ to match locations in LCSS. We compute the LCSS distance with $\epsilon=0.005$ and $\epsilon=0.02$ respectively. According to the Spearman's rank correlations between different distance measures listed in Table \ref{table:distance}, inconsistency exists between metrics. The correlation coefficients exceed 0.74 between the Cosine distance derived from GPT-J embeddings and the Hausdorff and DTW distances, while the LCSS distance is sensitive to the matching threshold with a higher dissimilarity.

\begin{table}[h]
\centering
\caption{The results of Spearman's rank correlations between different trajectory distance measures}
\begin{tabular}{|c|c|c|c|c|c|}
\hline
                    & \textbf{Hausdorff} & \textbf{DTW} & \textbf{LCSS($\epsilon=0.005$)} & \textbf{LCSS($\epsilon=0.02$)} & \textbf{Cosine} \\ \hline
\textbf{Hausdorff}  & 1.000              & 0.973        & 0.597              & 0.761               & 0.749           \\ \hline
\textbf{DTW}        & 0.973              & 1.000        & 0.596              & 0.755               & 0.747           \\ \hline
\textbf{LCSS($\epsilon=0.005$)}  & 0.597              & 0.596        & 1.000              & 0.799               & 0.482           \\ \hline
\textbf{LCSS($\epsilon=0.02$)} & 0.761              & 0.755        & 0.799              & 1.000               & 0.537           \\ \hline
\textbf{Cosine(GPT-J)}     & 0.749              & 0.747        & 0.482              & 0.537               & 1.000           \\ \hline
\end{tabular}
\label{table:distance}
\end{table}

\begin{figure}[!h]
    \centering
    \includegraphics[width=.8\linewidth]{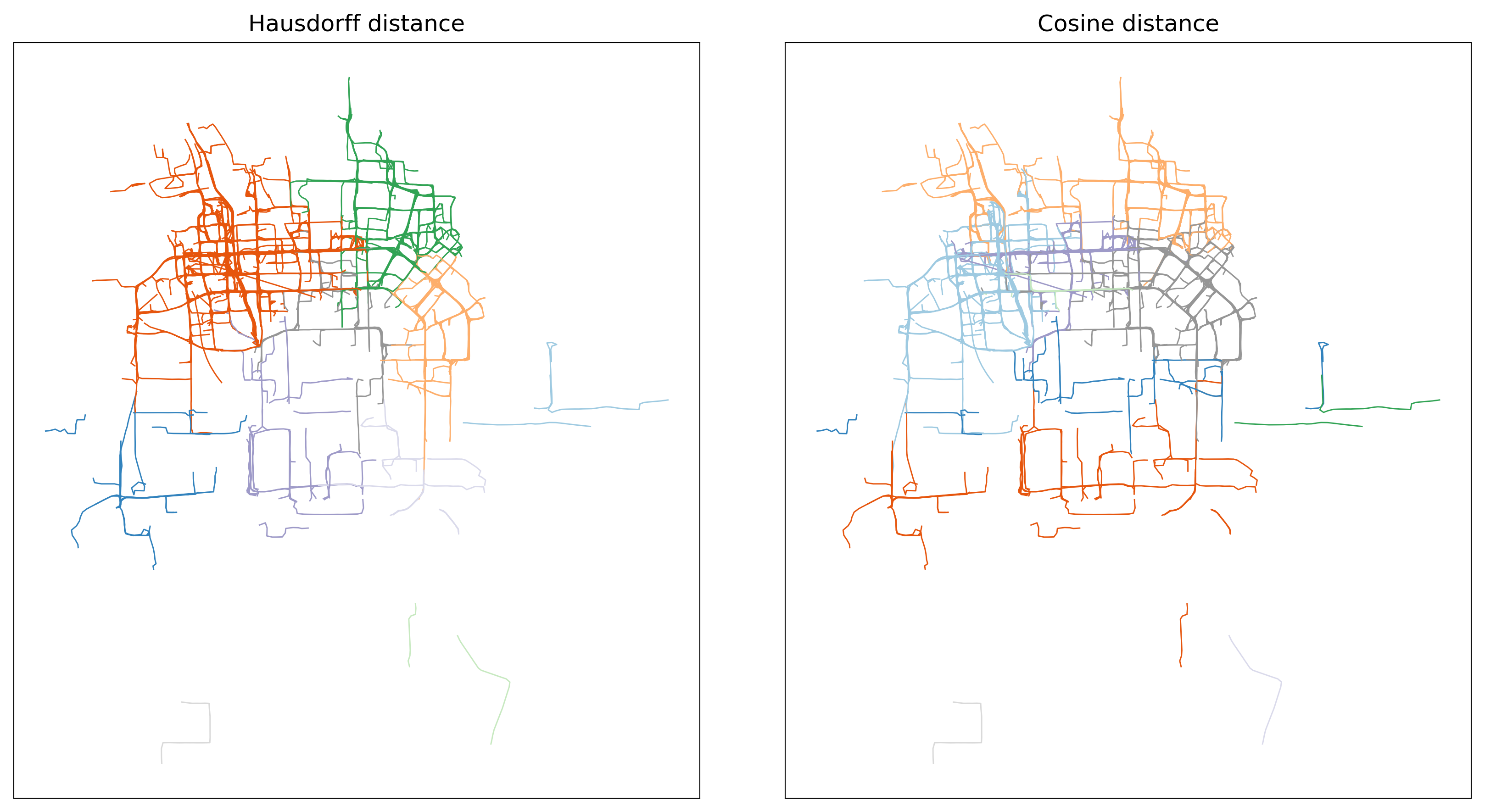}
    \caption{Trajectory clustering results of the same user using different distance measurements (color indicates different clusters)}
    \label{fig:clusters}
\end{figure}

Figure \ref{fig:clusters} displays the clustering results ($n\_clusters=10$) using the Hausdorff distance of raw trajectories and the Cosine distance of trajectory embeddings). The Rand Index of 0.68 suggests a substantial agreement and shared clusters between both methods. When considering the nearest neighbors, the overlapping neighbors are 36.8\% and 48.3\% when k=5 and 20 respectively, implying a discrepancy in the neighbors. Interestingly, despite differences in the characteristics of the neighbors and the shape and range of identified clusters, the trajectories clustered based on cosine distance exhibit spatial clustering patterns, highlighting the usability of the LLM embedding-based distance in trajectory clustering applications. 

\subsection{Trajectory destination prediction results} \label{results}

We filter the trajectories whose destination is located in the Haidian District, the city of Beijing. Then we truncate the trajectory to the last 15 minutes and extract the 75\% section of a trajectory along with its destination. The dataset of 6,298 trajectories is partitioned into three subsets, with an 80:5:15 split ratio for training, validation, and testing, respectively. Five sets of candidate destinations are generated for validation and training data. The results are shown in Table \ref{table:destination_error} and \ref{table:destination_acc}. After the training process, 97\% of the output from LLMs are tuned to be valid coordinates. In addition, regarding the quality of the predicted destination, the fine-tuned CausalLM model outperforms the GMM-based prediction when the trajectory clusters are set as 25. When we generate more candidate destinations, the overall error decreases, leading to a significantly increased accuracy of from 0.08 (Top-1 Accuracy) to 0.62 (Top-5 Accuracy) when the distance is relatively short as 100m. Moreover, if the distance is further loosen to 500m, the accuracy of Top-1 destination prediction can reach 0.57 while the Top-5 accuracy can even reach 0.86. The results demonstrate the capability of integrating the LLMs, which are primarily transformer-based architecture, to understand the spatiotemporal dependency contained in trajectories and perform location prediction tasks successfully.


\begin{table}[h]
\centering
\caption{Trajectory destination prediction results: validity and error (NA: not available)}
\begin{tabular}{|c|cc|cc|cc|}
\hline
\multirow{2}{*}{Method} & \multicolumn{2}{c|}{Validity@5}  & \multicolumn{2}{c|}{Error@1}     & \multicolumn{2}{c|}{Error@5}     \\ \cline{2-7} 
              & \multicolumn{1}{c|}{Valid} & Test & \multicolumn{1}{c|}{Valid} & Test & \multicolumn{1}{c|}{Valid} & Test \\ \hline
LLM: CausalLM           & \multicolumn{1}{c|}{0.97} & 0.97 & \multicolumn{1}{c|}{0.62} & 0.59 & \multicolumn{1}{c|}{0.58} & 0.58 \\ \hline
Baseline: GMM & \multicolumn{1}{c|}{1}     & 1    & \multicolumn{1}{c|}{1.29}  & 1.32 & \multicolumn{1}{c|}{NA}    & NA   \\ \hline
\end{tabular}
\label{table:destination_error}
\end{table}

\begin{table}[h]
\centering
\caption{Trajectory destination prediction results: accuracy (NA: not available)}
\begin{tabular}{|c|cc|cc|cc|cc|}
\hline
\multirow{2}{*}{Method} &
  \multicolumn{2}{c|}{\begin{tabular}[c]{@{}c@{}}Accuracy@1\\ (100m)\end{tabular}} &
  \multicolumn{2}{c|}{\begin{tabular}[c]{@{}c@{}}Accuracy@5\\ (100m)\end{tabular}} &
  \multicolumn{2}{c|}{\begin{tabular}[c]{@{}c@{}}Accuracy@1\\ (500m)\end{tabular}} &
  \multicolumn{2}{c|}{\begin{tabular}[c]{@{}c@{}}Accuracy@5\\ (500m)\end{tabular}} \\ \cline{2-9} 
              & \multicolumn{1}{c|}{Valid} & Test & \multicolumn{1}{c|}{Valid} & Test & \multicolumn{1}{c|}{Valid} & Test & \multicolumn{1}{c|}{Valid} & Test \\ \hline
LLM: CausalLM & \multicolumn{1}{c|}{0.08}  & 0.08 & \multicolumn{1}{c|}{0.62}  & 0.63 & \multicolumn{1}{c|}{0.55}  & 0.57 & \multicolumn{1}{c|}{0.83}  & 0.86 \\ \hline
Baseline: GMM & \multicolumn{1}{c|}{0.01}  & 0.01 & \multicolumn{1}{c|}{NA}    & NA   & \multicolumn{1}{c|}{0.10}  & 0.07 & \multicolumn{1}{c|}{NA}    & NA   \\ \hline
\end{tabular}
\label{table:destination_acc}
\end{table}

\section{Conclusion}
\label{sec:conclusion}

In this research, we utilize the pretained GPT-J  foundation model to encode the string format of trajectories and then evaluate the effectiveness of LLM-based embeddings for trajectory data analysis in the era of GeoAI. The experiments demonstrate that while the LLM-based embeddings can preserve certain trajectory distance metrics, challenges remain in estimating numeric values and retrieving spatial neighbors for destination prediction. In addition, the LLMs can understand the spatiotemporal dependency contained in trajectories and has good accuracy in location prediction task. Future work will focus on evaluation of other LLMs (e.g., GPT-4, GPT-4o, Llama 3, and Claude 3) and the exploration of semantic queries on trajectory data with attributes (e.g., activity mode, trip purpose, place visits). This research highlights the need for improvement in terms of capturing the nuances and complexities of the underlying geospatial data and integrating domain knowledge to support various GeoAI applications using LLMs. Looking ahead, when we integrate more AI foundation models into the human movement data analytics, we also need to pay attention to the increasing concerns regarding the location privacy, biases, and ethical issues in LLMs ~\cite{rao2023building,xie2023geo}.

\bibliography{references}

\begin{thebibliography}{10}

\bibitem{besse2017destination}
P.~C. Besse, B.~Guillouet, J.-M. Loubes, and F.~Royer.
\newblock Destination prediction by trajectory distribution-based model.
\newblock {\em IEEE Transactions on Intelligent Transportation Systems},
  19(8):2470--2481, 2017.

\bibitem{demvsar2021establishing}
U.~Dem{\v{s}}ar, J.~A. Long, F.~Benitez-Paez, V.~Brum~Bastos, S.~Marion,
  G.~Martin, S.~Sekuli{\'c}, K.~Smolak, B.~Zein, and K.~Si{\l}a-Nowicka.
\newblock Establishing the integrated science of movement: bringing together
  concepts and methods from animal and human movement analysis.
\newblock {\em International Journal of Geographical Information Science},
  35(7):1273--1308, 2021.

\bibitem{dodge2020progress}
S.~Dodge, S.~Gao, M.~Tomko, and R.~Weibel.
\newblock Progress in computational movement analysis--towards movement data
  science.
\newblock {\em International Journal of Geographical Information Science},
  34(12):2395--2400, 2020.

\bibitem{gao2023handbook}
S.~Gao, Y.~Hu, and W.~Li.
\newblock {\em Handbook of Geospatial Artificial Intelligence}.
\newblock CRC Press, 2023.

\bibitem{janowicz2020geoai}
K.~Janowicz, S.~Gao, G.~McKenzie, Y.~Hu, and B.~Bhaduri.
\newblock {GeoAI: spatially explicit artificial intelligence techniques for
  geographic knowledge discovery and beyond}.
\newblock {\em International Journal of Geographical Information Science},
  34(4):625--636, 2020.

\bibitem{kasneci2023chatgpt}
E.~Kasneci, K.~Se{\ss}ler, S.~K{\"u}chemann, M.~Bannert, D.~Dementieva,
  F.~Fischer, U.~Gasser, G.~Groh, S.~G{\"u}nnemann, E.~H{\"u}llermeier, et~al.
\newblock Chatgpt for good? on opportunities and challenges of large language
  models for education.
\newblock {\em Learning and Individual Differences}, 103:102274, 2023.

\bibitem{krumm2006real}
J.~Krumm.
\newblock Real time destination prediction based on efficient routes.
\newblock In {\em Society of Automotive Engineers (SAE) 2006 World Congress},
  volume~7. Citeseer, 2006.

\bibitem{mai2023opportunities}
G.~Mai, W.~Huang, J.~Sun, S.~Song, D.~Mishra, N.~Liu, S.~Gao, T.~Liu, G.~Cong,
  Y.~Hu, et~al.
\newblock {On the Opportunities and Challenges of Foundation Models for GeoAI
  (Vision Paper)}.
\newblock {\em ACM Transactions on Spatial Algorithms and Systems}, pages
  1--44, 2024.

\bibitem{mooney2023towards}
P.~Mooney, W.~Cui, B.~Guan, and L.~Juh{\'a}sz.
\newblock {Towards understanding the geospatial skills of ChatGPT: Taking a
  geographic information systems (GIS) exam}.
\newblock In {\em Proceedings of the 6th ACM SIGSPATIAL International Workshop
  on AI for Geographic Knowledge Discovery}, pages 85--94, 2023.

\bibitem{JSSv103i04}
L.~Pappalardo, F.~Simini, G.~Barlacchi, and R.~Pellungrini.
\newblock scikit-mobility: A python library for the analysis, generation, and
  risk assessment of mobility data.
\newblock {\em Journal of Statistical Software}, 103(1):1–38, 2022.

\bibitem{rao2023building}
J.~Rao, S.~Gao, G.~Mai, and K.~Janowicz.
\newblock Building privacy-preserving and secure geospatial artificial
  intelligence foundation models (vision paper).
\newblock In {\em Proceedings of the 31st ACM International Conference on
  Advances in Geographic Information Systems}, pages 1--4, 2023.

\bibitem{song2010limits}
C.~Song, Z.~Qu, N.~Blumm, and A.-L. Barab{\'a}si.
\newblock Limits of predictability in human mobility.
\newblock {\em Science}, 327(5968):1018--1021, 2010.

\bibitem{tsiligkaridis2020personalized}
A.~Tsiligkaridis, J.~Zhang, H.~Taguchi, and D.~Nikovski.
\newblock Personalized destination prediction using transformers in a
  contextless data setting.
\newblock In {\em 2020 International Joint Conference on Neural Networks
  (IJCNN)}, pages 1--7. IEEE, 2020.

\bibitem{gpt-j}
B.~Wang and A.~Komatsuzaki.
\newblock {GPT-J-6B: A 6 Billion Parameter Autoregressive Language Model}.
\newblock \url{https://github.com/kingoflolz/mesh-transformer-jax}, May 2021.

\bibitem{wang2020generalizing}
Y.~Wang, Q.~Yao, J.~T. Kwok, and L.~M. Ni.
\newblock Generalizing from a few examples: A survey on few-shot learning.
\newblock {\em ACM computing surveys (csur)}, 53(3):1--34, 2020.

\bibitem{xie2023geo}
Y.~Xie, Z.~Wang, G.~Mai, Y.~Li, X.~Jia, S.~Gao, and S.~Wang.
\newblock Geo-foundation models: Reality, gaps and opportunities.
\newblock In {\em Proceedings of the 31st ACM International Conference on
  Advances in Geographic Information Systems}, pages 1--4, 2023.

\bibitem{yang2022large}
X.~Yang, A.~Chen, N.~PourNejatian, H.~C. Shin, K.~E. Smith, C.~Parisien,
  C.~Compas, C.~Martin, A.~B. Costa, M.~G. Flores, et~al.
\newblock A large language model for electronic health records.
\newblock {\em npj Digital Medicine}, 5(1):194, 2022.

\bibitem{zheng2023chatgpt}
O.~Zheng, M.~Abdel-Aty, D.~Wang, Z.~Wang, and S.~Ding.
\newblock Chatgpt is on the horizon: Could a large language model be all we
  need for intelligent transportation?
\newblock {\em arXiv preprint arXiv:2303.05382}, 2023.

\bibitem{zheng2010geolife}
Y.~Zheng, X.~Xie, W.-Y. Ma, et~al.
\newblock Geolife: A collaborative social networking service among user,
  location and trajectory.
\newblock {\em IEEE Data Eng. Bull.}, 33(2):32--39, 2010.

\bibitem{zheng2011computing}
Y.~Zheng and X.~Zhou.
\newblock {\em Computing with spatial trajectories}.
\newblock Springer Science \& Bussiness Media, 2011.

\end{thebibliography}

\end{document}